\documentclass[11pt]{article}

\usepackage{PRIMEarxiv}

\usepackage[utf8]{inputenc} 
\usepackage[T1]{fontenc}    
\usepackage{hyperref}       
\usepackage{url}            
\usepackage{booktabs}       
\usepackage{amsfonts}       
\usepackage{nicefrac}       
\usepackage{microtype}      
\usepackage{lipsum}
\usepackage{fancyhdr}       
\usepackage{graphicx}       
\graphicspath{{media/}}     

\usepackage{xcolor}
\usepackage{float}
\usepackage{array}
\usepackage{multirow}

\newcommand\MyBox[2]{
  \fbox{\lower0.75cm
    \vbox to 1.7cm{\vfil
      \hbox to 1.7cm{\hfil\parbox{1.4cm}{#1\\#2}\hfil}
      \vfil}%
  }%
}

\title{UID as a Guiding Metric for Automated Authorship Obfuscation\footnote{This research was conducted at Pennslyvania State University through an NSF funded REU. The entire research and paper creation process took eleven weeks, starting in late May of 2023 and ending early August 2023.}
}

\author{
  Nicholas Abegg \\
  Saint Vincent College \\
  \texttt{nicholas.abegg@stvincent.edu} \\
}

\begin{document}
\maketitle

\begin{abstract}
Protecting the anonymity of authors has become a difficult task given the rise of automated authorship attributors. These attributors are capable of attributing the author of a text amongst a pool of authors with great accuracy. In order to counter the rise of these automated attributors, there has also been a rise of automated obfuscators. These obfuscators are capable of taking some text, perturbing the text in some manner, and, if successful, deceive an automated attributor in misattributing the wrong author. We devised three novel authorship obfuscation methods that utilized a Psycho-linguistic theory known as Uniform Information Density (UID) theory. This theory states that humans evenly distribute information amongst speech or text so as to maximize efficiency. Utilizing this theory in our three obfuscation methods, we attempted to see how successfully we could deceive two separate attributors. Obfuscating 50 human and 50 GPT-3 generated articles from the TuringBench dataset, we observed how well each method did on deceiving the attributors. While the quality of the obfuscation in terms of semantic preservation and sensical changes was high, we were not able to find any evidence to indicate UID was a viable guiding metric for obfuscation. However, due to restrictions in time we were unable to test a large enough sample of article or tune the parameters for our attributors to comment conclusively on UID in obfuscation.
\end{abstract}


\section{Introduction}
With recent significant advances in Artificial Intelligence (AI), it has become possible for AI methods to automate the process of authorship attribution (AA). AA is the process of taking a text of unknown authorship and attributing the authorship amongst a number of known authors. These automated AA methods are capable of attributing an author with great accuracy (\cite{Uchendu2020-nk}). As a result, anonymity for authors has become endangered, and their ability to release controversial text is in question. Adversarially to AAs, there has been a rise in AI-driven authorship obfuscators (AO), which are automated methods that take some text and obscure the original author through some perturbation. AOs are successful if it is capable of deceiving an AA into attributing the incorrect author of a text. AOs also need to manipulate the text in a way that the ideas and messages originally conveyed in the text are preserved as well as having any changes to the text be subtle enough for human investigators not to notice.
There are consistently new developments for novel AA and AO methods, often with each trying to counter vulnerabilities exposed by recent developments in its countering process. As a result of this adversarial relationship, developing novel AO or AA methods is crucial for improving both tasks. In this paper, we explore a novel obfuscation method that is capable of obfuscating a text utilizing Uniform Information Density (UID) metrics as a guiding measure. UID is a psycho-linguistic theory that suggests that humans distribute information evenly throughout a message so as to maximize efficiency and clarity. Not only do humans follow these UID patterns, but it has been shown that language models follow their own respective UID pattern that is separate from humans (\cite{Venkatraman2023-bx}). We hypothesize that we could utilize UID as a guiding metric in the obfuscation of an article that results in a automated AA misattributing an obfuscated article. The alteration of authorship classification as a result of our obfuscation technique could indicate UID as a promising metric to be further investigated for the task of AO. Currently, no other obfuscation method utilizes UID theory, which means that exploring its performance and applicability could yield further improvements in AO techniques, and eventually improvements in AA as well.

\subsection{Research Question} 
\paragraph{RQ:}Explore the applicability of UID metrics in the task of obfuscation. Specifically, investigate if UID can be used as a guiding metric to result in successful obfuscation in which an automated authorship attributor misattributes an obfuscated article.

\hfill \break To explore this problem, we designed three algorithms known as Synonym Swap, UID Word Swap (UWS) and UID Paraphrase (UP). Synonym Swap and UWS involves selecting a word roughly in each sentence and swapping it out for a suitable synonym amongst a list of candidate synonyms. UP involves using diversity beam search for word replacement based on UID metrics. We will explore the design, implementation, and results of all three algorithms later in this paper.

\subsubsection{Limitations: Addressing Scope}
It is important to note that due to restrictions on time, this research will mainly focus on investigating the application of our novel AO technique using a small number of articles from the TuringBench dataset. Therefore, it is essential to consider that our results and findings could be subject to errors and warrant further investigation and exploration to draw more definitive and concrete conclusions about the efficacy of UID as a metric for obfuscation.

Another limitation we faced was the lack of accurate and easy to use detectors for the articles. Our first experiments, those done on Synonym Swap, used OpenAI's detector which was taken down on July 20, 2023. As a result of the detector being removed, we were forced to utilize two open-source implementations of models Detect GPT and ZeroGPT out-of-the-box with no tuning.

\subsection{Literature Overview}
\subsubsection{Authorship Attribution
}
The ability to attribute an author of some text to a specific author amongst a pool of authors has been accomplished in a multitude of ways. With the rise of machine generated text, there have been AA models that are capable of achieving near-perfect accuracy in identifying the language model (LM) behind a given text (\cite{Munir2021-yt}). There are also methods of deobfuscation, which is the process of reducing the impact obfuscation processes have on AA's ability to successfully attribute the correct author (\cite{Zhai2022-pc}).  Yet, the quality and sophistication of modern LMs mean they are capable of producing text of human-level quality that confuses machine authorship classifiers (\cite{Uchendu2020-nk}). There have also been successful efforts at AA for code. One particular study identified differentiating characteristics in programming style between ChatGPT and human-generated code. This allowed for significantly accurate classification of code authorship between human-authored and ChatGPT generated code (\cite{Ke2023-nl}).
For a more comprehensive overview of the state of AA and its literature, there are two recent and in-depth surveys (\cite{Uchendu2022-in}) (\cite{Zheng2023-pn})

\subsubsection{Authorship Obfuscation}
\paragraph{Rule-Based Obfuscation}
The task of authorship obfuscation is typically approached in two ways. There are rules-based methods in which rule(s) or guideline(s) are set and the method attempts to alter the text through some perturbation that results in the rule(s) or guideline(s) being satisfied. An example of rule based obfuscation is utilizing a general stylometric average of specific stylometric characteristics such as punctuation, word frequency, part of speech frequency, etc., to push an author's stylometric metrics in line with these averages. The hope would be that the changes push an author's text towards this general average and would obfuscate an author's identifiable style and preserve their anonymity (\cite{Karadjov2017-el}). Another rule-based method employed an approach very similar to the methods we devised for this paper. Observing the words that appeared most frequently in an author's text, the obfuscation method attempted to select a synonym to replace the word in order to mask an author’s word choice and frequency stylometric signature. Using WordNet to generate these synonyms and calculating the Wu and Palmer similarity score of each synonym to find the most suitable replacement for a given word. This resulted in obfuscated text that still retained much of the quality of the original text as at most only one word per sentence was changed (\cite{Mansoorizadeh_undated-rf}).
A notable novel obfuscation technique takes a heuristic approach to the AO task for the first time. The approach involves modeling style using Jensen-Shannon distance. Then obfuscating the text through a heuristic paraphrasing operation that greedily maximizes obfuscation at each step. Specifically, aiming to distance the style of the obfuscated text from the author's style and move towards a more generic style \cite{bevendorff-etal-2019-heuristic}.
\paragraph{Internal AA Obfuscation}
The other approach involves accessing an AA classifier so as to guide the task of obfuscation. Methods such as Mutant-X (\cite{Mahmood2019-wt}), A4NT (\cite{Shetty2017-vz}), and ParChoice (\cite{Grondahl2019-iv}) utilize a classifier to ensure successful obfuscation. Mutant-X assumes black-box knowledge, which means it has access to the adversary’s trained classifier as well as its input features for reference. Mutant-X involves employing a fitness function which considers both the attribution probability from the adversary classifier as well as semantic similarity. A notable advantage is that it requires no text previously written by the author for training, making it significantly easier to use (\cite{Mahmood2019-wt}). A4NT obfuscates an author's text by imitating specific target classes, such as converting a female-authored text to emulate a male-authored text instead. The imitation is achieved by referencing the target classifier to guide the changes that would result in a male classification instead (\cite{Shetty2017-vz}). Parchoice takes a piece of text, paraphrases it to generate multiple paraphrase candidates, and then selects the most appealing candidate. For document-level obfuscation, the method is rather similar to Mutant-X, as ParChoice paraphrases a random sentence in the document and selects the candidate that would be most likely to cause misclassification by the classifier (\cite{Grondahl2019-iv}).
\subsection{Psycho-linguistics and Uniform Information Density}
Uniform Information Density (UID) theory suggests that humans optimize their speech and text so as to uniformly distribute information over an entire message (\cite{Frank2008}). Essentially humans tend to prefer communicating with others in a way that spreads information throughout the sentence in a balanced manner. Furthermore, information content in sentences can be quantified so as to measure and represent the UID of different messages (\cite{Meister2021-tu}). Recently, it has also been found that not only do humans demonstrate and follow particular patterns for UID, but decoding algorithms follow their own patterns as well (\cite{Venkatraman2023-bx}). The ability to to quantify UID and the fact that both humans and machines demonstrate unique patterns of UID, indicate the possibility that UID could be a useful metric in emulating the authorship of specific LMs or a human author.

\section{Preliminary \& Framework}
\subsection{TuringBench Dataset}
The dataset we utilized is the TuringBench dataset which consists of 200k news articles with 20 labels. One label is human, which represents human generated articles from media outlets such as CNN, The Washington Post, and Breitbart. The 19 others are different neural text-generators such as GPT-3, GROVER, FAIR, etc. that generate articles utilizing headlines as the generation prompts.(\cite{Uchendu2021-ed}). The dataset is specifically designed for the “Turing Test” problem, which involves being able to differentiate machine generated text and human generated text. This makes the dataset a perfect source for longer form article content that are pre-labeled allowing us to compare the classification labels to the ground truth labels.

\subsection{Semantic Cosine Similarity Score}
To select a proper alternate article for obfuscation, we needed to calculate the semantic similarity score of the new articles produced by UID Word Swap. We utilized Scikit Learn's Pairwise submodule cosine similarity function to calculate the cosine similarity between an obfuscated article and the original unaltered article. We decided to compare the semantic similarity of entire articles rather than individual sentences.

\subsection{UID Score}
Uniform Information Density (UID) is defined as the preference for humans to distribute information within a sentence uniformly, maximizing efficiency in communication \cite{Meister2021-tu}. To select a proper candidate, we needed to calculate the UID score of each probable obfuscated article to determine which resulted in the greatest difference of UID from the original. For all articles, we calculated four different UID metrics and selected the two that seemed to have the most variation. The two UID metrics we decided to utilize were the UID score variance and UID score ${difference}^2$. To calculate both of these scores, we first computed the surprisal of the tokens in the article. Surprisal is defined as the negative log probability of a token given the previous tokens in the context of the article. The UID score $variance$ is obtained by calculating the $variance$ of the surprisals for all tokens within the article. To obtain the UID score ${difference}2$, we calculate the average of the squared differences in surprisals for every two consecutive tokens within the article.

\subsection{Authorship Attributors}
To test how our obfuscation methods affect an automated authorship attributor, we utilized two attributors, ZeroGPT(\cite{AITextDetector}) and DetectGPT(\cite{mitchell2023detectgpt}), to test the effects on our data. Out first experiment utilized OpenAI's detector which was retired as of July 20, 2023.

\section{Synonym Swap: A Simple Test of Obfuscation}
\subsection{Design and Implementation Overview}
Prior to the creation of the UWS algorithm, we designed and tested a more primitive obfuscation technique that we called Synonym Swap. Synonym Swap centered around taking some number of articles and attempting to obfuscate each sentence of each article. The obfuscation technique consisted of swapping out a target word within a sentence for a synonym that was calculated to be the most probable next token by the GPT-2 language model. To select the target word, Synonym Swap would start in the middle of a sentence and check each word from there to the end of the sentence, until it found the first suitable candidate. The criteria for selecting a suitable candidate consists of the word not being a stop word, having at least one synonym, not being labeled as a proper noun, and has more than 2 characters. Stop words are highly frequent words that are avoided as target words as they lack significance within a sentence concerning informational content. Once a suitable candidate has been selected, Synonym Swap pools all of the synonyms for the target word using the WordNet corpus from NLTK and calculates which one is most probable to appear as the next token in the sentence given the previous tokens of the sentence. Once the most probable synonym replacement has been selected, the sentence is then altered to contain this new word and the process is repeated until the end of the article. If a sentence was too short or did not have a suitable candidate then it is left untouched by Synonym Swap.
\subsection{Experiments}
To test the effectiveness of Synonym Swap we used 100 articles from the TuringBench dataset, with 50 being human articles and the other 50 being generated by GPT-3. Each article obfuscated by Synonym Swap and then classified by a binary Authorship Attributor, OpenAI Detector\footnote{As of July 20, 2023, OpenAI's detector has been taken down due to poor accuracy in detection.}. The classifier takes a piece of text and generates the probability that a given text is written by a machine as well as assigning a label to the article. There are 5 different possible labels that the classifier can assign to a piece of text: very unlikely, unlikely, unclear if it is, possibly, or likely.
We had the classifier label both the original 50 human and 50 GPT-3 generated texts and the Synonym Swap obfuscated texts to compare the effect that the obfuscation had on the classifier.
\subsection{Results}
The results indicated that the simple Synonym Swap was capable of confusing the classifier and causing it to generate different labels as a result of the obfuscation.
Furthermore, the simple Synonym Swap was capable of producing semantically similar and reasonable sentences, but would also generate unreasonable or nonsensical changes to the text. Often these unreasonable results were a result of WordNet, which was caused by it grabbing all possible synonyms for a word which often did not fit the specific context in which the swap was occurring. For example, when targeting the word “may” in the context such as “May I leave the room?,” synonyms generated would include words related to the month of May and various kinds of flowers.

\subsection{Graphs for Synonym Swap}
\begin{figure}[H]
    \centering
    \includegraphics[width=.75\linewidth]{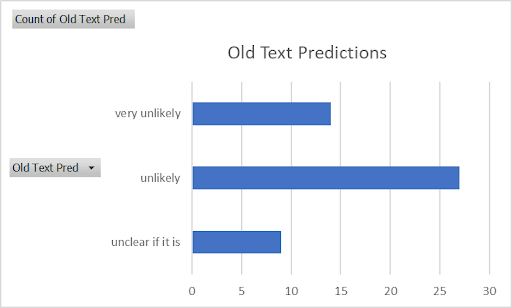}
    \caption{Unaltered Human: Synonym Swap Labels for Original Results (unlikely and very unlikely refer to if the text was generated by an AI)}
    \label{UnalteredHuman}
\end{figure}

\begin{figure}[H]
    \centering
    \includegraphics[width=.75\linewidth]{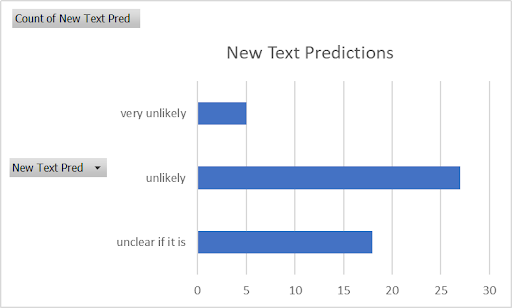}
    \caption{Altered Human: Synonym Swap Labels for Original Results (unlikely and very unlikely refer to if the text was generated by an AI)}
    \label{AlteredHuman}
\end{figure}

\begin{figure}[H]
    \centering
    \includegraphics[width=.75\linewidth]{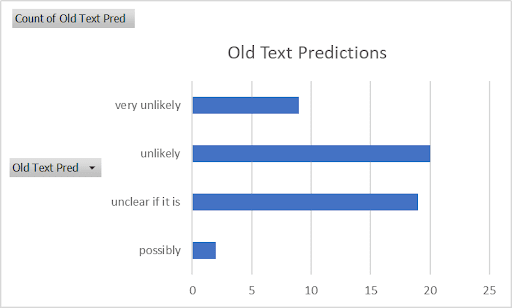}
    \caption{Unaltered Machine: Synonym Swap Labels for Alternate Article Results (unlikely and very unlikely refer to if the text was generated by an AI)}
    \label{UnalteredMachine}
\end{figure}

\begin{figure}[H]
    \centering
    \includegraphics[width=.75\linewidth]{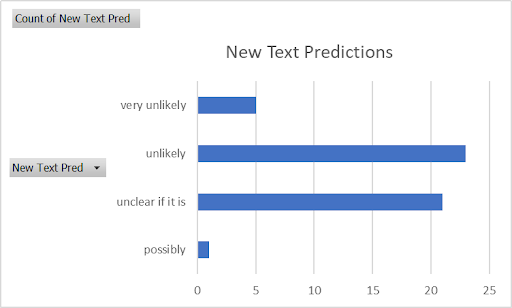}
    \caption{Altered Machine: Synonym Swap Labels for Alternate Article Results (unlikely and very unlikely refer to if the text was generated by an AI)}
    \label{AlteredMachine}
\end{figure}

These plots provided us with some evidence that even a simple of obfuscation like Synonym Swap was capable of altering the labels produced by the detector. Specifically, Figure 1 and Figure 2 show a drastic change in the articles marked very unlikely AI generated. It seems our changes made it more unclear to the detector on whether the text was AI generated or not. Figure 3 and 4 also show a similar pattern of making it more unclear to the detector if a text is AI generated or not. While it is a very small experiment, with only 100 articles being used, it still indicates some promise to the method of obfuscation via single word swapping per sentence.

\subsection{Text Generation Quality}
Below we provide some samples of the text generated by synonym swap compared to the original articles.
\paragraph{Human Generated Text (Previous Label: Very Unlikely AI generated, New Label: Unclear if it is)}

\hfill \break \textbf{Human Text (Original):} “a look at some of donald trump's early activity as \textcolor{red}{president}: -- 24: executive orders and memoranda signed. that includes to withdraw the nited states from trade deal, impose federal hiring \textcolor{red}{freeze} reduce regulations related health care law enacted under former president barack obama. 1: blocked. an order ban travelers…”

\hfill \break \textbf{Synonym Swap Text:} "a look at some of donald trump's early activity as \textcolor{red}{President\_of\_the\_United\_States} -- 24: executive orders and memoranda signed. that includes to withdraw the nited states from trade deal, impose federal hiring \textcolor{red}{stop\_dead} reduce regulations related health care law enacted under former president barack obama. 1: hinder an order ban travelers…”

\hfill \break The above examples showcase the text quality generated by Synonym Swap on a human generated piece of text which was able to successfully deceive the AA after its changes. The red colored text are the target words with the top part of the text representing the original and the bottom part showing the obfuscated version. As shown in the text, the quality of the Synonym Swap alterations can range from acceptable to wholly unreasonable. The first swap, "president" to "President\_of\_the\_United\_States" is an example of a swap that is satisfactory. An example of a incorrect swap that Synonym Swap can make is the swap "freeze" to "stop\_dead". While the phrase is a synonym to freeze, it fails to fit the context of the sentence and is considered a incorrect swap.

\paragraph{Machine Generated Text (Previous Label: Very Unlikely AI generated, New Label: Unlikely AI generated)}
\hfill \break \textbf{Machine Text:} ”and tensions in the middle east?curtis' heartachethe .k. is sadly one of leading countries dealing with coronavirus, as \textcolor{red}{another} 14 cases have been confirmed now country. total, there 20 incidents april 4, 2014, none it's ended death.on itv's show this morning, gloria hunnisett talks a parent who lost their daughter at 21 days old due to virus. curtis' story haunting absolutely heartbreaking.they didn't think was anything \textcolor{red}{wrong} her until few later when she admitted hospital.”
\hfill \break \textbf{Synonym Swap Text:} “and tensions in the middle east?curtis' heartachethe .k. is sadly one of leading countries dealing with coronavirus, as \textcolor{red}{some\_other} 14 cases have been confirmed now country. total, there 20 incidents april 4, 2014, none it's ended death.on itv's show this morn gloria hunnisett talks a parent who lost their daughter at 21 days old due to virus. curtis' story haunting absolutely heartbreaking.they didn't think was anything \textcolor{red}{haywire} her until few later when she admitted hospital.”

In the above example, the swaps showcase more of synonym swaps capabilities to make changes to text. The first swap, "another" to "some\_other" is reasonable and correct swap that fits within the context of the sentence. However, the second swap, "wrong" to "haywire", is entirely unreasonable and completely fails to capture the original idea and content of the original sentence.

\subsection{Discussion}
Synonym Swap allowed us to get an idea of the effectiveness of simply swapping single words in a sentence on an automated authorship attributor. Our results indicated that even a technique that randomly swaps words with no logical process behind its target selection can still influence an AA’s labels on the text. It also informed us of the need to implement a new method for selecting replacement words for obfuscation. For Synonym Swap, we were limited to the Wordnet corpus and its ability to generate likely synonyms for a target word, which often proved insufficient. We also relied on GPT-2’s ability to select the most probable word from these selected synonyms, given only the previous tokens of the current sentence. Thus, it was unable to consider any information contained in the sentence after its location had been determined, which further limited Synonym Swap’s ability to preserve semantics.
With these lessons in mind, we decided to design a new obfuscation technique to correct many of these faults as well as further enhance its ability to select target candidates in sentences more logically. This technique, UID Word Swap (UWS), utilizes DistilBERT, a compressed version of BERT, instead of GPT-2. DistilBERT is a large language model that is specifically trained to excel at masked language modeling which is precisely the task that we are undertaking when swapping a word to obfuscate the text. Not only does DistilBERT fix our problem with only having the context of the sentence up to and not after the target word, but it also removes the need for using the Wordnet corpus. Since DistilBERT is specifically trained for masked language modeling, it is capable of selecting semantically similar and reasonable words without referring to a corpus such as Wordnet.

\section{UID Word Swap \& Paraphrase}
Following our preliminary algorithm, Synonym Swap, we developed two more algorithims for obfuscation. The first is UWS, which utilizes DistilBERT to generate the probable words in a masked language model format. The second is UID Paraphrase, which uses diverse beam search to generate diverse paraphrased versions of the articles. Both methods generated 10 new articles for each of the 100 articles in our small data set. We will go into further specific details about the implementation and results for both next.

\subsection{UID Word Swap}
\subsubsection{Implementation and Design}
UID Word Swap (UWS) is a further iteration on Synonym Swap that was designed from the ground up with the goal of improving on the aspects in which Synonym Swap failed. Mainly, poor word choice in relation to the context of the sentence. Using DistilBERT to generate the candidate words for a particular target word, UWS not only generates higher quality results that are more semantically similar, but also does so significantly faster. Below we will provide a pseudo code algorithm of UWS to demonstrate the general process of the algorithm.
\newline
\noindent\rule{12cm}{0.4pt}
\begin{verbatim}
UID_Word_Swap(masked_sentence)
{
    sentence_tokens = tokenizer(masked_sentence)

    sentence_logits = model(sentence_tokens)

    # get 10 most probable tokens by sorting 10 highest scored tokens
    most_prob_tokens = sort(logits)[-10:]

    most_prob_words = decode(most_prob_tokens)
    for each prob_word in most_prob_words
    {
        new_sentence = masked_sentence.replace([MASK], prob_word)
        most_prob_sentences_list.append(new_sentence)
    }
    return most_prob_sentences_list
}

parse_sentences(article, num_of_article)
{
    article_sentences = sentence_tokenize(article)

    for each sentence in article_sentences
    {
        sentence_words = word_tokenize(article_sentences)

        part_of_speech = get_pos(sentence_words)

        if sentence_words >= 3 words
        {
            target_word = middle word in the sentence
            target_index = middle word index in sentence
        }
        if target_index is not null
        {
            while target_word is not a stop word or punctuation  mark, does not have synonyms, 
            is not alphabetical, or is a proper noun
            {
                target_word = next word in sentence
                target_index = next word index
            }
            masked_sentence = sentence
            masked_sentence[target_index] = [MASK] token

            probable_sentences.append(UID_Word_Swap(masked_sentence))
        }
        else
        {
            probable_sentences.append(None)
        }
    }
    return probable_sentences
}

human_gpt3_articles = 50 human + 50 gpt3 from TuringBench

for each article in human_gpt3_articles
{
    parse_sentences(article, number_of_article)
}
\end{verbatim}

\noindent\rule{12cm}{0.4pt}

\hfill \break
As shown in the pseudo code above, we utilized a masked language modeling approach to conduct the word swap. Specifically, on lines 351-357 is where the target word is determined and then the [MASK] token is inserted into the sentence. Notice the target word criteria is very similar to the target word criteria of Synonym Swap. WordNet is even queried to check if a word has synonyms in an effort to only pick words that would be easier to swap. At this point, the masked sentence is given to DistilBERT which then determines the ten most probable words for the swap. Ten alternate sentences are constructed for every sentence in which a suitable word was located. These ten alternate sentences are then used to construct ten new articles with each article being made up of the ith item in the alternate sentence list constructed for each sentence in the article. For the sentences in which no suitable word was located, the original sentence is used for all the ten alternate articles.

\paragraph{Candidate Selection Criteria:} Following the creation of these ten alternate articles, we needed to implement the UID score and semantic similarity scores as a guiding metric. We did this by simply calculating the UID $variance$ and ${Difference}^2$ for the original articles as well as their ten alternates. Due to the semantic similarity scores being very high for UWS obfuscated articles, we set the required threshold for similarity to be $>= 0.98$ or 98\% similar to the original. The scores are so high as a result of the changes to each article being at most one word swap per sentence in the article. For the two UID scores, we calculated the difference for each alternate article's UID scores, compared to the original's UID scores. For both $variance$ and ${Difference}^2$, the articles with the largest UID differences to the original were inspected until one was found that satisfied the semantic similarity score threshold. The plots for demonstrating the UID vs Semantic Similarity Scores are shown in the following results section.

Below is a pseudo code outline of the candidate selection process used for the UWS algorithm. 
\hypertarget{pseudo}{any sentence}
\newline
\noindent\rule{12cm}{0.4pt}
\begin{verbatim}
candidate_select(num_article)
{
    original_article = the original article
    alternate_list = ten alternate articles generated by UWS 
    
    original_UID1 = the original UID $variance$ score
    alternate_UID1 = the ten alternate UID $variance$ scores
    
    original_UID2 = the original UID difference squared score
    alternate_UID2 = the ten alternate UID difference squared scores
    
    for each alternate article's UID $variance$ Score
        append the difference of UID $variance$ score and original UID $variance$ score
            
    for each alternate article's UID difference squared Score
        append the difference of UID difference squared score and
        original UID difference squared score
        
    sorted_UID1 = sorted(UID1_Difference_list)
    sorted_UID2 = sorted(UID2_Difference_list)
    
    for each Variacne and Difference Squared score in sorted_UID1 and sorted_UID2
        get semantic similarity score of current alternate article
        if semantic similarity >= .98:
            select given article
            break
}
\end{verbatim}

\noindent\rule{12cm}{0.4pt}

\subsubsection{UID Word Swap Results}
Below we will present the results of the experiments we ran to test the effectivness and capabilities of the UWS obfuscation process.
\subsubsection*{Scatter Plots UID vs. Semantic Similarity}
\begin{figure}[H]
    \centering
    \includegraphics[width=1\linewidth]{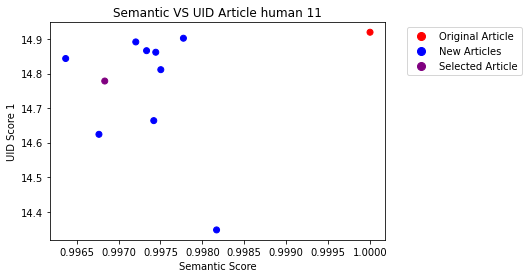}
    \caption{UID Word Swap utilizing UID Score ($variance$)}
    \label{UWS Figure}
\end{figure}
\begin{figure}[H]
    \centering
    \includegraphics[width=1\linewidth]{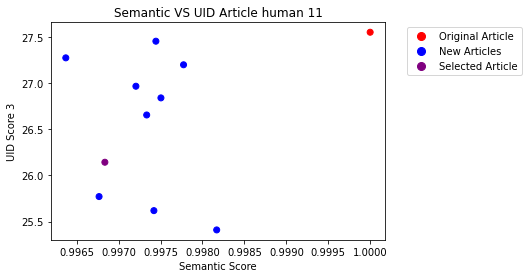}
    \caption{UID Word Swap utilizing UID Score (${Difference}^2$)}
    \label{UWS Figure 2}
\end{figure}

To showcase our results of UWS's candidate selection, we generated two scatter plots for each article that mapped the correlation of semantic similarity score (x-axis) with the UID scores (y-axis) between the original article and its ten alternative articles.The first scatter plot represents the UID Score 1 ($variance$), while the second scatter plot corresponds to UID Score 3 (${difference}^2$).
Notice that the Semantic Score of UWS is rather high. This is due to the fact that only one word in each sentence at most is being swapped. This indicates higher levels of semantic preservation, which is expected given that the change to each article is rather small. This also seems to result in little, but still noticeable, difference across the UID scores for each article. Notice the original article, marked in red, has a semantic similarity score of 1.0 as a result of being compared to itself.

\paragraph{Text Generation Quality}
UWS’s results proved to generate more consistent and higher quality sentences. While Synonym Swap only produced one new alternate sentence, which is then generated into a new article, UWS generated 10 new sentences for each sentence it was given. These 10 new sentences were then used to generate 10 new alternate articles. Each of these articles then had a semantic similarity score and four UID scores calculated. These scores were then used to generate a plot for each article, which allows us to see which article is most semantically similar while also being significantly different with their UID score differences.
However, due to UWS only altering one word per sentence in each article, all articles had very similar UID scores. To address this issue, we developed another algorithm known as UID Paraphrase that utilizes Diversity Beam Search to generate more diverse obfuscated articles.

The plots for the sample text below can be found above, figure \ref{UWS Figure} and \ref{UWS Figure 2}.
\paragraph{Human Generated Text:}
 \hfill \break \textbf{Original Article (Human):} “washington ( ) at a time when president donald trump seems to permeate nearly every aspect of american discourse, it might come as surprise that the first movie from barack and michelle obama's \textcolor{red}{production} company, higher ground, never mentions him by name.but subtlety is part power factory, new netflix documentary charts r ening factory in dayton, ohio. over course two hours, movie, directed seasoned documentarians steven bognar julia reichert, serves quiet historical political corrective, offering their portrait state america's industrial heartland \textcolor{red}{prodding} viewers rethink who, exactly, project.american starts on december 23, 2008, crowd gathers learn general motors plant dayton has shuttered. then fast-forwards 2015, ens enterprise, fuyao glass \textcolor{red}{america}, arm shanghai-based company manufactures automotive glass. one man makes fuyao's expanded mission crystal clear: what we're doing melding cultures together: chinese culture s culture. so we are truly global organization.as some critics have pointed out, is, important ways, commentary \textcolor{red}{unpredictability} globalization; york times review frames underscoring haves have-nots.but it's also much more than that. arrives moment white house continues make vociferous, bold claims about economy, particularly manufacturing. that's despite increasing concerns economists warnings history recession could be horizon. there's sobering contrast between trump's rhetoric how job growth ballooned during his \textcolor{red}{presidency} reality broader slowdown slamming states -- including ohio helped win 2016 presidential election.read”
 
 \hfill \break \textbf{UID Word Swap Text UID Score $variance$ \& ${difference}^2$:} “ washington ( ) at a time when president donald trump seems to permeate nearly every aspect of american discourse , it might come as surprise that the first movie from barack and michelle obama 's \textcolor{red}{publishing} company , higher ground , never mentions him by name.but subtlety is part power factory , new netflix documentary charts r ening factory in dayton , ohio . over course two hours , movie , directed seasoned documentarians steven bognar julia reichert , serves quiet historical political corrective , offering their portrait state america 's industrial heartland \textcolor{red}{helps} viewers rethink who , exactly , project.american starts on december 23 , 2008 , crowd gathers learn general motors plant dayton has shuttered . then fast-forwards 2015 , ens enterprise , fuyao glass \textcolor{red}{corporation} , arm shanghai-based company manufactures automotive glass . one man makes fuyao 's expanded mission crystal clear : what we 're doing bringing cultures together : chinese culture s culture . so we are truly global organization.as some critics have pointed out , is , important ways , commentary \textcolor{red}{upon} globalization ; york times review frames underscoring haves have-nots.but it 's also much more than that . arrives moment white house continues make vociferous , conflicting claims about economy , particularly manufacturing . that 's despite increasing concerns economists thought history recession could be horizon . there 's sobering contrast between trump 's rhetoric how job growth ballooned during his \textcolor{red}{political} reality broader slowdown slamming states -- including ohio helped win 2016 presidential election.read”

 \hfill \break Notice here some of the successful and unsuccessful word swaps that occurred.  The first swap, production being swapped for publishing, is a successful swap as the meaning is preserved. Another example of a successful swap is prodding being swapped for helps. However, an example of a slightly unsuccessful is the swapping of "america" for corporation since this is altering a proper noun, which is something that should be avoided as it alters the semantics of the sentence in a significant way.

\paragraph{Machine Generated Text:}
\hfill \break \textbf{Machine Text (GPT-3:} “'vatican city -- a tribunal of senior clerics at the vatican is about to deliver verdict in one biggest sex abuse cases it has \textcolor{red}{faced} since start catholic church's scandal.the level hypocrisy simply appalling, vincenzo folena, an italian editorialist, networked cbs. like osorno's, he said, church sanctions fellow human beings employing cruel tactics, sending them exile and punishment places for something which they do not have feel guilty--this whilst abusers that hid behind osorno's cassock will never appear before court.osorno's scandal conflicts?osorno's appointed bishop, juan barros, been \textcolor{red}{accused} snapping after priest, father fernando karadima, was found guilty ruling launched 2011.bishop barros produced nation's osorno 2015, pope francis engineered his placement; delegation prominent catholics subsequent unsuccessfully tried pit appointment down.pope defended decision promote 2015 trip chile.the archbishop santiago, cardinal ricardo ezzati, decorated by ? 2016. likewise walked global holocaust memorial day \textcolor{red}{events}, although jewish community offended church'”

\hfill \break \textbf{UID Word Swap Text UID Score $variance$ \& difference squared:} “ 'vatican city -- a tribunal of senior clerics at the vatican is about to deliver verdict in one biggest sex abuse cases it has \textcolor{red}{convened} since start catholic church 's scandal.the level hypocrisy simply appalling , vincenzo folena , an italian editorialist , networked cbs . like osorno 's , he said , church sanctions fellow human beings employing cruel tactics , sending them exile and punishment places for something which they do not have feel guilty -- this whilst abusers that hid behind osorno 's cassock will never appear before court.osorno 's scandal conflicts ? osorno 's appointed bishop , juan barros , been \textcolor{red}{investigated} snapping after priest , father fernando karadima , was found guilty ruling launched 2011.bishop barros produced nation 's osorno 2015 , pope francis engineered his placement ; delegation prominent catholics subsequent unsuccessfully tried pit appointment down.pope defended decision promote 2015 trip chile.the archbishop santiago , cardinal ricardo ezzati , decorated by ? 2016. likewise walked global holocaust memorial day \textcolor{red}{weekend} , although jewish community offended church '”

\hfill \break Here we can see some successful and unsuccessful alterations to a GPT-3 generated text. The first swap of faced being swapped for convened is rather unsuccessful and fails to preserve the semantic idea of the original article. A successful swap is shown with the swapping of "accused" for "investigated". This is deemed sensible and semantically similar swap.

\subsection{UID Paraphrase}
\subsubsection{Implementation \& Design}
UID Paraphrase (UP) builds upon UWS and Synonym Swap but changes the process of obfuscation by paraphrasing each sentence instead of swapping just a word. To select if a sentence is eligible for obfuscation we ensured the sentence was equal to or longer than 8 characters. The sentence is then tokenized and entered into a diverse beam search model that attempts to generate new sentences that are semantically similar.
The diverse beam search model decodes the sentence and considers all possible next tokens that could appear in the sentence, scoring each with the T-5 Large Paraphraser Language model. Utilizing the diversity penalty hyper parameter, we encourage the model to produce more diverse outputs, so as to get more semantically unique articles. The idea is that these unique articles are diverse enough to yield higher variation in their UID scores while still maintaining a relatively high semantic similarity to the original.
Following a system similar to UWS, we generate ten alternate sentences that are then used to generate the ten new articles. From the list of ten alternate sentences for each sentence, the ith alternate sentence is picked and used to make the ith article.

\paragraph{Candidate Selection Criteria:} Similar to UWS, after the creation of the 10 alternate articles, we then calculate the semantic cosine similarity scores for each article compared to the original article, as well as UID score $variance$ and UID score ${Difference}^2$. The candidate selection process follows a similar method of sorting the difference in UID scores between the alternate articles and the original articles. We then check the highest difference in UID scores until one is found that has a semantic similarity score that passes the threshold. Due to the greater variation in the articles for UP than in UWS, we changed the threshold for semantic similarity to $>= 0.85$ to accommodate the increased diversity in the alternate articles. The candidate selection is shown in pseudo code \hyperlink{pseudo}{here}.

\subsubsection{UID Paraphrase Results}
\paragraph{Scatter Plots UID vs. Semantic Score}
\hfill \break

\begin{figure}[H]
    \centering
    \includegraphics[width=1\linewidth]{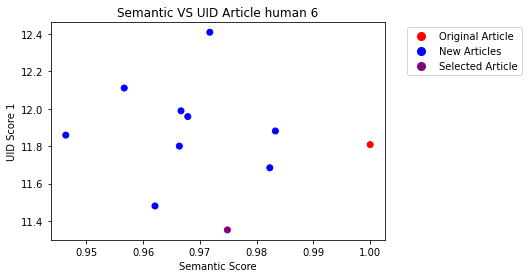}
    \caption{UID Paraphrase Swap utilizing UID Score ($variance$)}
    \label{UID1A6}
\end{figure}

\begin{figure}[H]
    \centering
    \includegraphics[width=1\linewidth]{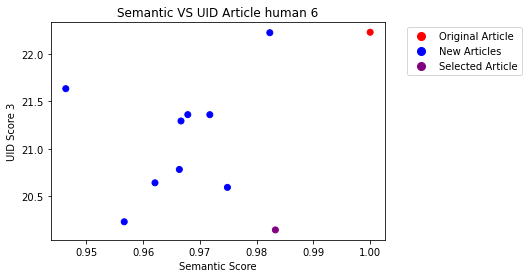}
    \caption{UID Paraphrase Swap utilizing UID Score (${Difference}^2$)}
    \label{UID3A6}
\end{figure}

The scatterplots help illustrate the process of selecting a proper candidate. For this particular example, we can observe the selected articles (colored purple) are furthest from the UID score of the original article, yet they still satisfy the semantic similarity threshold.

\paragraph{Text Generation Quality}
Below is an example of a human generated piece of text and its resulting UID paraphrased selected articles. Notice this particular article has two separate articles selected based on the different $variance$ and ${difference}^2$ UID scores.
\paragraph{Human Generated Text:}
\hfill \break \textbf{Human Text (Original):} “the nited nations has ended a campaign featuring wonder woman as an ambassador for women and girls, two months after announcement was met with protests petition complaining that fictional superhero inappropriate choice to represent female empowerment. in announcing october, said it about girls everywhere, who are their own right, men boys support struggle gender equality. but not everyone saw way. nearly 45, 000 p le signed protesting selection. a white of impossible proportions, scantily clad shimmery, body suit american flag motif boots is appropriate spokeswoman equity at nations, said. jeffrey brez, spokesman .”

\hfill \break \textbf{UID Paraphrase UID Score variance:} “A campaign involving wonder woman as an ambassador for women and girls has ended two months after being met with protests petition claiming that the fictional superhero inappropriate choice to represent female liberation.  Girls everywhere, who are their own right, and men boys support gender discrimination are all in announcing october.. but not everyone saw way.  More than half of the population of nearly 45, 000 p les have been rejected in the protesting vote.  According to a white of insignificant proportions, scantily clad shimmery, body suit american flag motif boots, is appropriate spokeswoman equity in nations.. jeffrey brez, spokesman .”

\hfill \break \textbf{UID Paraphrase UID Score ${Difference}^2$:} “Two months after being met with protests petition claiming that the fictional superhero inappropriate choice to represent female liberation, the nited nations has ended a campaign starring wonder woman as an ambassador for women and girls.  Girls everywhere, who are their own right, and men boys support gender discrimination are all in announcing october. but not everyone saw way.  Over 55,000 people were rejected in the protesting vote, with nearly 45, 000 p le signed protesting selection.  According to the author, a white of insignificant proportions, scantily clad, body suit american flag motif boots, is appropriate spokeswoman equity in nations. jeffrey brez, spokesman . ”

\subsection{Authorship Attributor Labeling Results:}
We then took the selected and original articles for both UWS and UP and obtained labels for each article. We utilized two automated authorship attributors, Zero GPT and DetectGPT, to obtain the labels. The labels were used to compare the effectiveness of UWS and UP in obfuscating the authorship of the articles. The detectors seemed to fair poorly at detecting machine generated text and seemed to label almost all of the articles human, even if they were GPT-3.

\paragraph{Note on Results:} It is important to notice that the results below are all calculated with both UID score results combined into a list and then compared to their original articles. This is why there are 200 altered articles as there are 100 $variance$ and 100 ${difference}^2$ articles and 200 original articles for each article respectively.

\begin{figure}[H]
    \centering
    \includegraphics[width=1\linewidth]{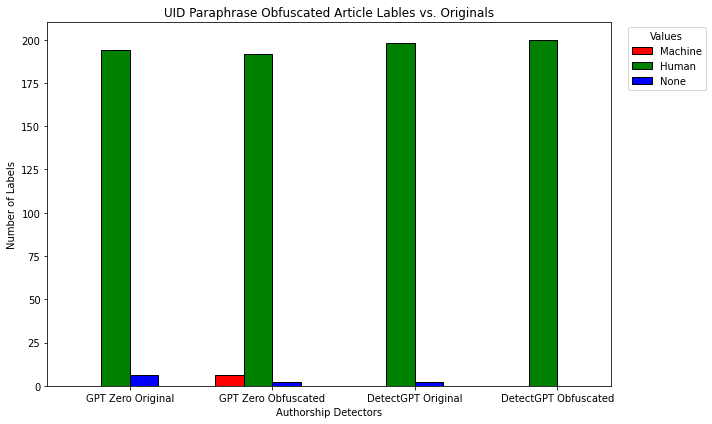}
    \caption{UID Paraphrase Labels}
    \label{ParaphraseLabel}
\end{figure}

\begin{figure}[H]
    \centering
    \includegraphics[width=1\linewidth]{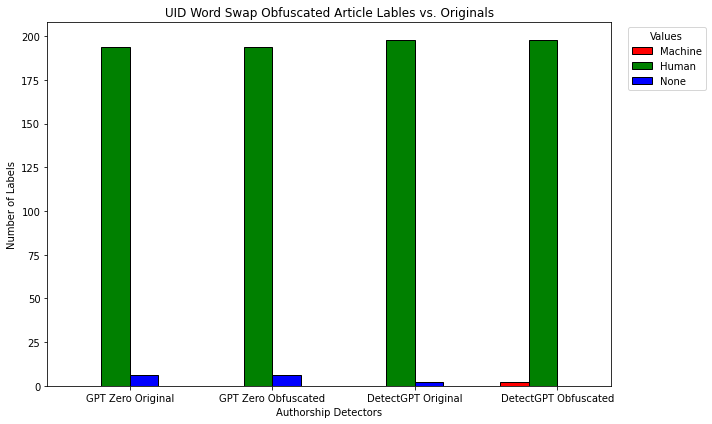}
    \caption{UID Word Swap Labels}
    \label{UIDWSLabel}
\end{figure}

\noindent
\renewcommand\arraystretch{1.5}
\setlength\tabcolsep{0pt}
\begin{tabular}{c >{\bfseries}r @{\hspace{0.7em}}c @{\hspace{0.4em}}c @{\hspace{0.7em}}l}
  \multirow{10}{*}{\parbox{1.1cm}{\bfseries\raggedleft actual\\ value}} & 
    & \multicolumn{2}{c}{\bfseries UID Word Swap GPT Zero} & \\
  & & \bfseries p & \bfseries n & \bfseries total \\
  & p$'$ & \MyBox{True}{Positive 4} & \MyBox{False}{Negative 96} & P$'$ \\[2.4em]
  & n$'$ & \MyBox{False}{Positive 2} & \MyBox{True}{Negative 98} & N$'$ \\
  & total & P & N &
\end{tabular}

\noindent
\renewcommand\arraystretch{1.5}
\setlength\tabcolsep{0pt}
\begin{tabular}{c >{\bfseries}r @{\hspace{0.7em}}c @{\hspace{0.4em}}c @{\hspace{0.7em}}l}
  \multirow{10}{*}{\parbox{1.1cm}{\bfseries\raggedleft actual\\ value}} & 
    & \multicolumn{2}{c}{\bfseries UID Word Swap DetectGPT} & \\
  & & \bfseries p & \bfseries n & \bfseries total \\
  & p$'$ & \MyBox{True}{Positive 2} & \MyBox{False}{Negative 98} & P$'$ \\[2.4em]
  & n$'$ & \MyBox{False}{Positive 0} & \MyBox{True}{Negative 100} & N$'$ \\
  & total & P & N &
\end{tabular}

The two confusion matrices above demonstrate the quality of the two detectors on the UWS results. Notice the split of false negatives and true negatives is nearly 50-50. This is due to the heavy favoring of the human label by the detectors.

Below are the F1 scores for paraphrase swaps. 

\begin{table}[htbp][H]
  \centering
  \caption{UID Paraphrase F1 Scores for GPT Zero and DetectGPT}
  \begin{tabular}{@{}lcc@{}}
    \textbf{Model} & \textbf{F1 Score} \\
    \midrule
    GPT Zero & 0.54 \\
    DetectGPT & 0.5 \\
    \bottomrule
  \end{tabular}
\end{table}

\noindent
\renewcommand\arraystretch{1.5}
\setlength\tabcolsep{0pt}
\begin{tabular}{c >{\bfseries}r @{\hspace{0.7em}}c @{\hspace{0.4em}}c @{\hspace{0.7em}}l}
  \multirow{10}{*}{\parbox{1.1cm}{\bfseries\raggedleft actual\\ value}} & 
    & \multicolumn{2}{c}{\bfseries UID Paraphrase Zero GPT} & \\
  & & \bfseries p & \bfseries n & \bfseries total \\
  & p$'$ & \MyBox{True}{Positive 8} & \MyBox{False}{Negative 92} & P$'$ \\[2.4em]
  & n$'$ & \MyBox{False}{Positive 0} & \MyBox{True}{Negative 100} & N$'$ \\
  & total & P & N &
\end{tabular}

\noindent
\renewcommand\arraystretch{1.5}
\setlength\tabcolsep{0pt}
\begin{tabular}{c >{\bfseries}r @{\hspace{0.7em}}c @{\hspace{0.4em}}c @{\hspace{0.7em}}l}
  \multirow{10}{*}{\parbox{1.1cm}{\bfseries\raggedleft actual\\ value}} & 
    & \multicolumn{2}{c}{\bfseries UID Paraphrase DetectGPT} & \\
  & & \bfseries p & \bfseries n & \bfseries total \\
  & p$'$ & \MyBox{True}{Positive 0} & \MyBox{False}{Negative 100} & P$'$ \\[2.4em]
  & n$'$ & \MyBox{False}{Positive 0} & \MyBox{True}{Negative 100} & N$'$ \\
  & total & P & N &
\end{tabular}

Once again, there is a similar split of false and true negatives that is due to the favoring of the human label by the detectors.

\begin{table}[htbp][H]
  \centering
  \caption{UID Word Swap F1 Scores for GPT Zero and DetectGPT}
  \begin{tabular}{@{}lcc@{}}
    \textbf{Model} & \textbf{F1 Score} \\
    \midrule
    GPT Zero & 0.51 \\
    DetectGPT & 0.51 \\
    \bottomrule
  \end{tabular}
\end{table}

Below is one of the examples where the attributor's labels between the two articles were different between the original label and the selected label. It seems that the diverse beam search had some problem generating a proper paraphrased article, leading to the nonsensical output shown in the selected article. It is possible that due to the beginning of the selected article being a sensical paraphrase of the original, it retained a high enough semantic similarity score. At the same time, the nonsensical ending of the paraphrased article resulted in a substantially different UID score, causing its score to be drastically different from the original's. Both of these events led to its selection as the paraphrased article. The nonsensical ending of the article is then likely the reason why the attributor altered its label to machine.

\paragraph{Original Article:}'the plan for mandatory debt-market defaulting in the life of dealer during first year rbi may postpone its to make compulsory all defaults debt market when it is launched, chief economic adviser arvind subramanian today said he told reporters that not be feasible have default one yearmarket expectations rs 50 rate cut soon; raghuram rajan's wish list yet revealed: shah - business standardreserve bank governor rajan india and philip clarke, ceo tesco k expectation a basis points next monetary policy meets with consensus, according standard report as growth known, participants eagerly wait decisionrbi set 1 2 2\% over five years: shaktikanta das linerbi line 's key variable repo rate, which displayed an increasing trend period 2011-14 declined 2015 following easing : reserve moneycontrol.comrbi displayed'

 \paragraph{UID Paraphrase Variance Selected Article:}According to a shakti bank governor's mandatory debt-market defaulting in the first year of the bank's life, the plan for mandatory debt-market defaulting in the life of the dealer during the first year rbi's first year may postponed, but the main variable repo rate, which increased trend rate in 2015, remains unchanged, according to reserve moneycontrol.comrbi's forecast, rising trend rate remains unchanged, with increasing trend rate declines \color{red} rajan rbi a rajan rajan rajan rajan rajan rbi rbi rajan rajan rajan rajan rbi rbi rbi's first year rbi rbi rbi rbi rbi rbi rbi rbi rbi rbi rbi rbi rbi rbi rbi rbi rbi rbi rbi rbi rbi rbi rbi rbi rbi rbi rbi rbi rbi rbi rbi rbi rbi rbi rbi rbi rbi rbi rbi rbi rbi rbi rbi rbi rbi rbi rbi rbi rbi rbi rbi rbi rbi rbi rbi rbi rbi rbi rbi rbi rbi rbi rbi rbi rbi rbi rbi rbi rbi rbi rbi rbi rbi rbi rbi rbi rbi rbi rbi rbi rbi \color{black}

\section{Conclusion \& Discussion}
From the results we gathered, there is no indication that UID is a metric that can be used in the process of obfuscation. A number of factors could be the reason for the results failing to provide any interesting evidence. These factors could be the poor accuracy of the detectors, using a rather small sample of 100 articles, and a lack of fine tuning for the parameters for UP's diverse beam search. The quality of text generated by both UWS and UP were rather promising and were capable of producing semantically similar results that had rather varied UID results.
\subsection{Future Work}
While our results failed to indicate much about the effectiveness of UID as a metric for obfuscation, further investigation is needed as much of our results can be deemed erroneous as a result of the poor detectors and small dataset size.

Future work would firstly involve the tuning of an AA that would provide accurate labels on the original articles. The AAs we used for our experiments in this paper failed to properly label the original articles. In fact, the AAs seemed heavily biased towards labeling nearly every article as human-generated. Thus, to make any conclusive remarks about the success and capabilities of both UWS, UP, and UID scores as a metric entirely would require further investigation. Tuning certain parameters for the current AAs, Detect GPT and ZeroGPT, or using a different detector could yield better results.

While UID is calculated for each article, it is not directly used within the obfuscation process. Further work on both UWS and UP in which we utilize the scores to guide the changes within the article could prove more successful. For instance, currently UWS uses DistilBERT to pick the ten most probable words for each sentence in the article. Ten articles are then created and then after the UID scores are calculated. Possibly involving the UID scores in the creation of the alternate articles or in the selection of the ten probable words could yield more impactful results on the labeling of the articles.

Other improvements we could make to our research would be as follows:
\begin{itemize}
    \item Increase number of articles as well as number of authors (other than human or GPT-3)
    \item Increase sophistication of both UWS and UP
    \item Further implement UID as a guiding metric within the obfuscation process
\end{itemize}

\section*{Acknowledgments}
I want to thank Dr. Dongwon Lee (Penn State University) for providing advice and guidance on conducting research. Furthermore, I would also like to thank Saranya Venkatraman (Penn State University \& New York University) who provided much advice and assistance while conducting the research. Saranya also conducted the test on the Authorship Attributors which provided me with the results used in this paper. This paper would not have been possible without the assistance of both Dr. Lee and Saranya.\footnote{This research was conducted at Pennslyvania State University through an NSF funded REU. The entire research and paper creation process took eleven weeks, starting in late May of 2023 and ending early August 2023.}

\bibliographystyle{unsrt}  
\bibliography{references}

\end{document}